\documentclass{article}

\PassOptionsToPackage{numbers, compress}{natbib}

\usepackage[preprint]{neurips_2026}

\usepackage[utf8]{inputenc} 
\usepackage[T1]{fontenc}    
\usepackage[pagebackref, breaklinks, colorlinks]{hyperref}
\usepackage{url}            
\usepackage{booktabs}       
\usepackage{amsfonts}       
\usepackage{nicefrac}       
\usepackage{microtype}      
\usepackage{xcolor}         
\usepackage{graphicx}       
\usepackage{multirow}
\usepackage{colortbl}
\usepackage{subcaption}
\usepackage{amsmath}
\usepackage{cleveref}
\usepackage{wrapfig}
\usepackage{siunitx}  
\usepackage{tabularx} 
\usepackage[table]{xcolor} 

\title{LVDrive: Latent Visual Representation Enhanced Vision-Language-Action Autonomous Driving Model}

\author{%
  Xiaodong Mei$^{1,}$\thanks{Work done when Xiaodong Mei was working as an intern at Xiaomi EV.} \\
  \And
  Diankun Zhang$^{2}$ \\
  \And
  Hongwei Xie$^{2,}$\thanks{Project leader.} \\
  \And
  Guang Chen$^{2}$ \\
  \AND
  Hangjun Ye$^{2}$ \\
  \And
  Dan Xu$^{1,}$\thanks{Dan Xu is the corresponding author.}
  \AND
  $^{1}$The Hong Kong University of Science and Technology, $^{2}$Xiaomi EV \\
  \texttt{xmeiab@connect.ust.hk, danxu@cse.ust.hk}
}

\begin{document}

\maketitle

\begin{abstract}
  
Vision-Language-Action (VLA) models have emerged as a promising framework for end-to-end autonomous driving. 
However, existing VLAs typically rely on sparse action supervision, which underutilizes their powerful scene understanding and reasoning capabilities. 
Recent attempts to incorporate dense visual supervision via world modeling often overemphasize pixel-level image reconstruction, neglecting semantically meaningful scene representation learning.
In this work, we propose \textbf{LVDrive}, a \textbf{L}atent \textbf{V}isual representation enhanced VLA framework for autonomous driving. 
LVDrive introduces a future scene prediction task into the VLA paradigm, where future representations are learned entirely in a high-level latent space under auxiliary supervision from a pretrained vision backbone. 
Departing from inefficient autoregressive generation, we jointly model future scene and motion prediction within a unified embedding space, processed in a single forward pass to conduct the future-aware reasoning. 
We further design a two-stage trajectory decoding strategy that explicitly leverages the learned latent future representations to refine trajectory generation.
Extensive experiments on the challenging Bench2Drive benchmark demonstrate that LVDrive achieves significant improvements in closed-loop driving performance, outperforming both action supervised methods and image-reconstruction-based world model approaches.
\end{abstract}

\begin{figure*}[h]
    \centering
    \includegraphics[width=0.99\textwidth, trim=0 360 0 0]{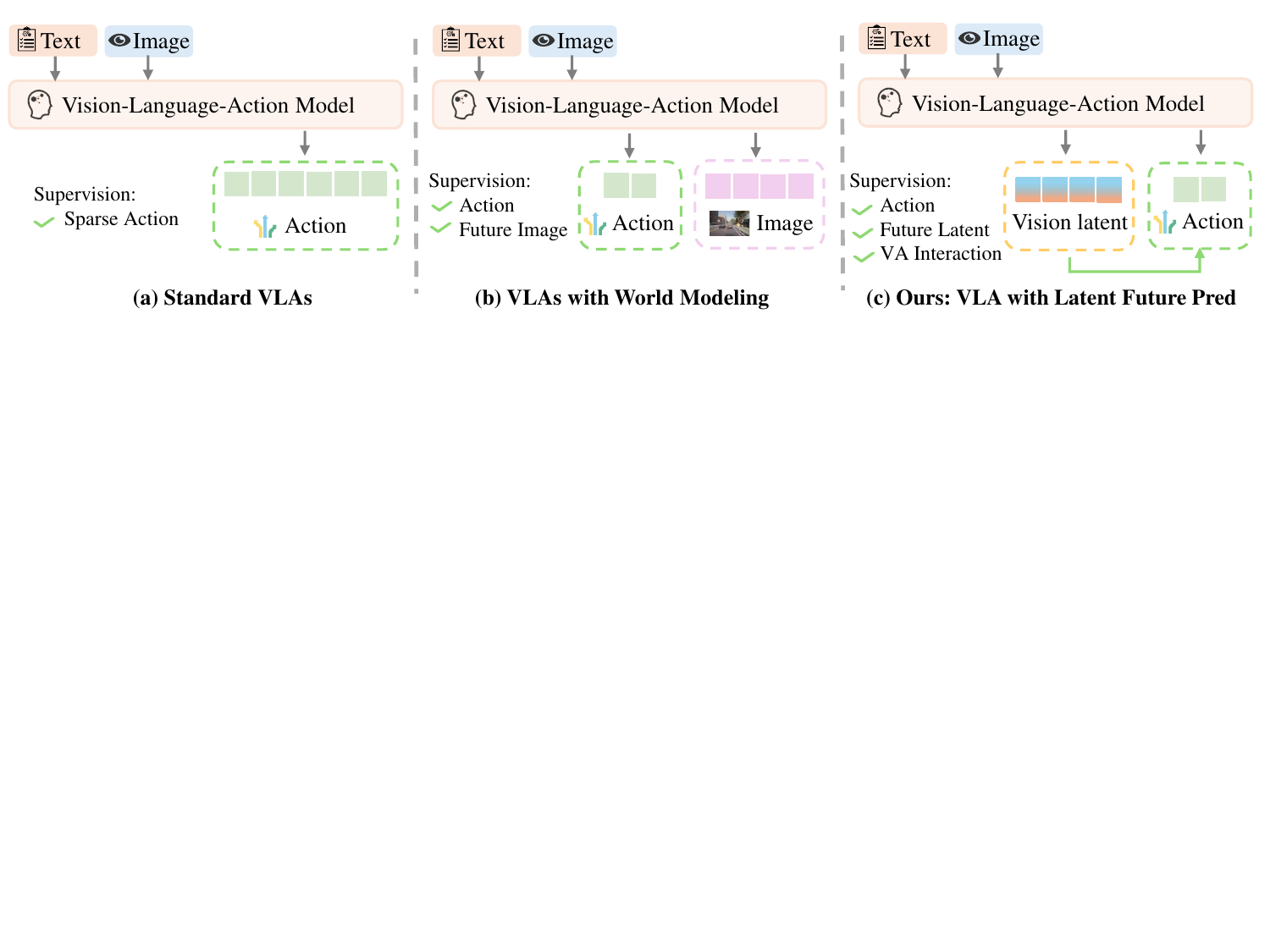}
    \caption{
    \textbf{The comparison of different VLA paradigms.}
    Standard VLA approaches, as depicted in (a), rely on sparse action supervision. Our LVDrive, illustrated in (c), performs the future visual and action representation learning jointly.
    Unlike VLA with the world modeling paradigm in (b), LVDrive predicts future scenes entirely in latent space, capturing rich semantic features without pixel-level reconstruction. The trajectory is then generated and explicitly refined through direct interaction with the learned future scene representations.}
    \vspace{-1em}
    \label{fig:intro}
\end{figure*}

\section{Introduction}

Leveraging the powerful scene understanding and reasoning abilities of multimodal foundation models, Vision-Language-Action (VLA) models~\cite{Hu2025VisionLanguageActionMF, Xu2023DriveGPT4IE, Hwang2024EMMAEM, Renz2025SimLingoVC, li2025recogdrive, Jiang2024SennaBL} hold immense potential for the end-to-end autonomous driving task, which can directly map raw sensor inputs and language instructions into future trajectories.
Current VLA models, however, suffer from a critical supervision deficit~\cite{Li2025DriveVLAW0WM, Jia2026DriveWorldVLAUL}: the exclusive reliance on sparse action labels leaves the structured spatial scene understanding and reasoning capacity of large foundation models substantially underutilized.

To enhance the spatial scene understanding capacity of large foundation models, recent frontiers have explored incorporating high-dimensional representations as auxiliary objectives, such as future visual feature prediction. 
Dense visual signals in the driving scenario exhibit a more natural alignment with ego-vehicle trajectories than language guidance alone, thereby grounding motion planning in the future-aware reasoning.
Inspired by the world modeling paradigm, existing approaches~\cite{Zeng2025FutureSightDriveTV, Xiong2026UniDriveWMUU, Li2025DriveVLAW0WM, Jia2026DriveWorldVLAUL, Zhou2025HERMESAU} learn to anticipate future visual observations for modeling environmental dynamics, thereby internalizing causal relationships that enhance planning performance.

Despite several efforts to exploit dense scene features, for end-to-end autonomous driving models with the ultimate objective of reliable motion planning, three persistent issues remain unaddressed in existing methods.
First, for a planning-oriented autonomous driving model, scene understanding should prioritize semantically rich representation learning over pixel-perfect frame reconstruction.
Overemphasizing texture-level reconstruction risks misleading the representation learning objectives, diverting the model's capacity away from high-level semantic understanding~\cite{Assran2023SelfSupervisedLF, Assran2025VJEPA2S}.
Second, formulating future frame and trajectory prediction as the next-token generation task with the discrete visual and action vocabulary is inherently time-consuming. 
The autoregressive formulation typically requires a large quantity of tokens to synthesize high-fidelity images and precise actions, which incurs prohibitive inference costs~\cite{Xiong2026UniDriveWMUU, Zheng2024Doe1CA, Li2025DriveVLAW0WM, Zhang2025EponaAD}.
Third, future visual features are consistently underutilized in the motion planning pipeline. Existing frameworks either confine future visual representations to implicit interactions with motion features~\cite{Xiong2026UniDriveWMUU, Zeng2025FutureSightDriveTV}, or employ future frames as a rudimentary world model to generate reward signals for trajectory optimization~\cite{Jia2026DriveWorldVLAUL}. Neither design is an explicit mechanism that embeds the rich semantic features directly into the trajectory generation.


To mitigate these challenges, we propose \textbf{LVDrive}, a framework that enhances VLA-based autonomous driving models with \textbf{L}atent \textbf{V}isual representation learning, shown in Figure~\ref{fig:intro}.
By leveraging dense future image frames as an additional supervision signal, our approach introduces a future scene prediction task into the VLA architecture.
Crucially, future scene representation is learned entirely within a latent space, without relying on texture-level reconstruction or hand-crafted annotations, so that the model is focused on capturing high-level semantic cues essential for motion planning.
Instead of autoregressively generating the discrete tokens, future scene features are learned alongside the trajectory decoding within the shared continuous optimization space in a single forward process, which naturally enhances the alignment between vision and action space.
With the future-aware reasoning, semantic scene features are gradually and implicitly injected into motion features, yielding a coarse predicted trajectory.
To fully unlock the benefits of predictive visual features, we further design a lightweight trajectory refiner that explicitly conditions on the learned future semantics to perform a fine-grained optimization of the trajectory decoding.
The two-stage design ensures that future scene information is not merely an auxiliary signal but an integral condition of the trajectory generation, leading to consistent improvements in closed-loop driving performance.

Our contributions are summarized as follows: 
\begin{itemize}
\item We introduce the latent future representation learning into the VLA-based autonomous driving framework, by jointly predicting the visual and action features in a single forward pass within a unified continuous space, to conduct the future-aware scene understanding and reasoning.
With the utilization of semantically rich visual features as auxiliary supervision, our approach strengthens the spatial scene understanding and reasoning capability of VLA model, which yields efficient and effective planning.
\item We design the two-stage trajectory decoding strategy to explicitly injects the learned future scene features into the motion generation.
Specifically, the coarse trajectory proposal is decoded after the future-aware reasoning, and further refined in a lightweight trajectory refiner module that conditions on the predicted semantic features, producing the more feasible and reasonable final trajectory.
\item Extensive experiments on the challenging Bench2Drive benchmark demonstrate that our method significantly improves the closed-loop driving performance, validating the effectiveness of dense future visual supervision exploitation in motion planning.
\end{itemize}

\section{Related Work}

\paragraph{VLAs for Autonomous Driving.}
Vision-Language-Action (VLA) models, built upon large multi-modal foundation models~\cite{liu2023llava, Qwen-VL, Beyer2024PaliGemmaAV}, have rapidly gained prominence in autonomous driving. 
Early approaches~\cite{Xu2023DriveGPT4IE, Ding2024HintADHA, Jiang2024SennaBL, Tian2024DriveVLMTC, Wang2023DriveMLMAM}
primarily employed VLMs as high-level semantic interpreters, generating scene descriptions and textual guidance to inform downstream planning performance.
More recent efforts have shifted toward end-to-end VLA frameworks~\cite{fu2025orion, Renz2025SimLingoVC, Zhou2025AutoVLAAV, Zhou2025OpenDriveVLATE, Renz2024CarLLaVAVL, Jiang2025DiffVLAVG, Hwang2024EMMAEM, Yang2025DriveMoEMF, Zeng2025FutureSightDriveTV, Zheng2024Doe1CA, li2025recogdrive, nvidia2025alpamayo, luo2026last, luo2026unleashing, liao2025diffusiondrive} that directly map raw sensor inputs and language instructions to driving actions, bypassing intermediate textual representations.
In the unified framework, actions are typically represented either as meta-action tokens drawn from the discrete vocabulary~\cite{Zhou2025AutoVLAAV, fu2025minddrive, tan2025latent} or as continuous motion features~\cite{fu2025orion, Yang2025DriveMoEMF, li2025recogdrive,nvidia2025alpamayo} that are subsequently decoded into the ego-trajectory through the action decoder.
Furthermore, the integration of reinforcement learning has recently been explored to boost the closed-loop performance of VLA driving models~\cite{fu2025minddrive, li2025recogdrive, nvidia2025alpamayo}.
Previous works~\cite{Li2025DriveVLAW0WM, Jia2026DriveWorldVLAUL} identify that the reliance of current VLAs on sparse action supervision and abstract language guidance underutilizes the spatial understanding capabilities of large multi-modal models. 
To this end, we propose a joint representation learning framework to incorporates auxiliary dense vision supervisions.

\paragraph{World Models for Autonomous Driving.}

World models, which aim to understand and model environmental dynamics, have emerged as a powerful paradigm for autonomous driving.
One line of research explores high-fidelity scene generation across various perceptual modalities, including images~\cite{Hu2023GAIA1AG, Zhang2025EponaAD, Gao2024VistaAG, Hassan2024GEMAG} and 3D representations~\cite{Xu2025OccLLMEA, Yang2023VisualPC, Zhang2023Copilot4DLU}, in order to construct the realistic simulator.
Another line seeks to integrate world modeling into end-to-end driving frameworks as a multi-task objective~\cite{Li2025DriveVLAW0WM, Xiong2026UniDriveWMUU, Zheng2024Doe1CA, liu2026driveva}, aiming to jointly enhance scene understanding and motion planning.
We argue that in this formulation, what matters for autonomous driving is not the fidelity of image synthesis, but rather the underlying world dynamics captured through predictive modeling, termed as latent world modeling~\cite{Ha2018WorldM}.
Several prior works~\cite{Li2024EnhancingEA, Zheng2025World4DriveEA, Lin2025FutureXEE} have explored leveraging latent world models within traditional end-to-end planning architectures.
Our work extends this idea to large VLA driving models, enabling the thorough and efficient exploitation of the reasoning capabilities inherent in large multi-modal foundation models.

\section{Methodology}
\label{method}
\begin{figure*}[t]
    \centering
    \includegraphics[width=0.99\textwidth, trim=0 260 90 0]{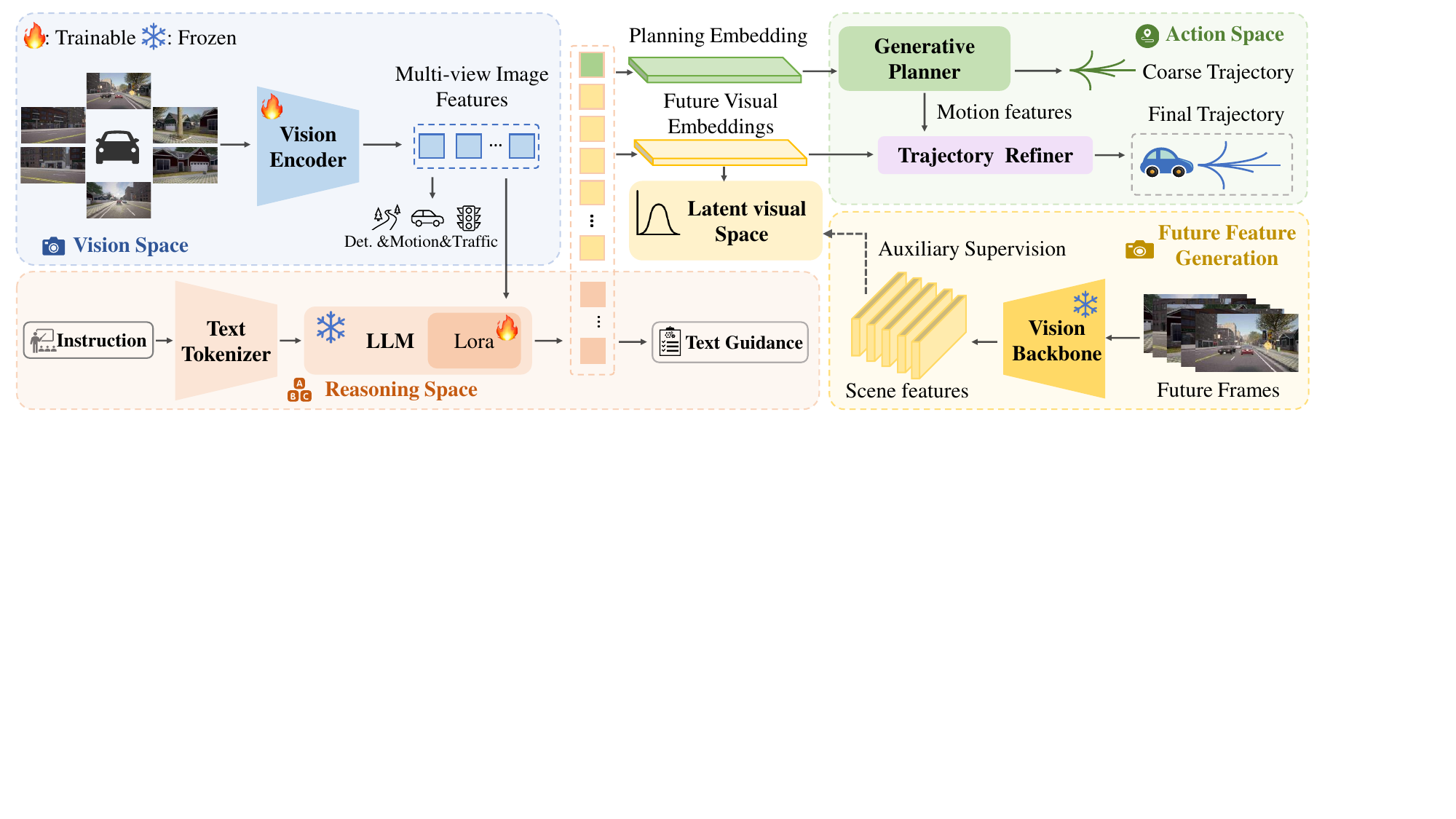}
    \vspace{-5pt}
    \caption{
    \textbf{Overview of LVDrive.} 
    LVDrive is a VLA framework that unifies latent future scene representation learning and motion planning, with dense auxiliary supervision provided by a pre-trained vision backbone. 
    Given multi-view images, the model encodes current and historical scene features and performs future-aware reasoning to predict both latent visual representations and motion features in a single forward pass. 
    The trajectory is then generated through a two-stage decoding process: a coarse proposal is first decoded from the planning embedding, and then refined by explicitly conditioning on the learned latent future scene features to produce the final trajectory.}
    \vspace{-15pt}
    \label{fig:model}
\end{figure*}

In this section, we present LVDrive, a VLA framework that jointly learns the future visual and action representations with the auxiliary semantic supervision signals from the pretrained vision backbone.
The model architecture is presented in Figure~\ref{fig:model}.
We incorporate the future scene prediction alongside the planning to generate the latent visual features in a single-forward process. The aligned vision and action representations initially enable the feasible trajectory generation. We further design the two-stage decoding strategy to produce the fine-grained trajectory conditioned on the latent vision feature explicitly.

\subsection{Problem Formulation}

Given a driving scenario at timestep $t$, we follow the recent end-to-end VLA model~\cite{fu2025orion}, where structured and historical multi-view image features $x_s, x_h$ are compressed and extracted through the dedicated vision encoder, and then projected with textual instructions $x_q$ into the unified token embedding space. 
To generate future continuous motion features in the time horizon $F$, the core LLM performs scene understanding and reasoning to produce the special planning token $s$, the embedding of which is subsequently decoded into the ego trajectory.  

We extend the task to predict the latent future visual features $V_{t+1:t+F}$ jointly with the planning, to enforce the model to conduct the future-aware reasoning. 
Motivated by the fact that human drivers attend primarily to the forward perspective, front-view prediction is sufficient for providing adequate future awareness.
Notably, the model relies solely on structured multi-view representations to predict future scene evolution.
This design effectively eliminates the potential noise introduced by texture-level details, allowing it to focus on learning semantically meaningful scene dynamics.
The whole task is formally formulated as

\begin{gather}
    V_{t+1:t+F}, s \sim p(V_{t+1:t+F},s|x_s, x_h, x_q). 
\end{gather}

\subsection{Latent Future Visual Representation Learning}
For the latent future visual feature prediction, we design a set of special latent visual placeholder tokens, termed \texttt{<img\_i>}, where $i \in \{0, 1,..., N\}$ indicates the number of tokens for each future image frame.
Special boundary markers are also designed as \texttt{<img\_start>} and \texttt{<img\_end>} to form the fixed output sequence of every future frame.
The sequence is repeated $F$ times to get multiple future scene predictions.

During the future-aware reasoning process, the model first locates all occurrences of latent visual placeholders in the outcomes, and divides them into multiple future frames at different timesteps according to the boundary markers.
Then the embedding of per-frame $H_{t+j} \in \mathbb{R}^{N \times D}$ is obtained in each placeholder span from the hidden states of the last LLM layer, where $D$ is the dimension of LLM and $j \in \{1,...,F\}$ denotes the number of future frames.
A lightweight vision decoder $\textsc{ViS}_{\theta}$ is subsequently applied to extract the semantic future features $V_{t+j}$:

\begin{gather}
     V_{t+j} = \textsc{ViS}_{\theta}(H_{t+j}), V_{t+j} \in \mathbb{R}^{N \times C_v}
\end{gather}
where $C_v$ is the dimension of latent visual feature.
The semantic supervision signals are derived from an independently pre-trained vision foundation model, which is maintained frozen in the training phase.
Its well-established scene understanding ability supplies the model with semantically meaningful guidance beyond the limited action labels and texture-level image reconstruction, contributing to more effective representation learning.

The special planning token \texttt{<waypoint\_ego>} is positioned after the latent visual sequence, allowing future scene representations to be implicitly aggregated into motion planning embedding via the causal attention mechanism.
The planning embedding $H_p \in \mathbb{R}^{1 \times D}$ is further decoded into the ego-trajectory through the two-stage trajectory decoding strategy, which is explicitly conditioned on latent future features.

\subsection{Two-stage Trajectory Decoding}
\label{two-stage decoding}

The ego-trajectory generation process are divided into two stages: to predict the coarse trajectory proposal $\tau_c \in \mathbb{R}^{K \times F \times 2}$ and then to generate the fine-grained final trajectory $\tau_f \in \mathbb{R}^{K \times F \times 2}$, with K indicating the number of multi-modal trajectories.

In the first stage, we utilize the VAE-based generative planner to produce the coarse proposals, following the designs in VAD~\cite{Jiang2023VADVS}, GenAD~\cite{Zheng2024GenADGE} and ORION~\cite{fu2025orion}.
Specifically, the planning embedding $H_p$ is mapped to the Gaussian distribution $p(z|H_p) \sim N(\mu, \sigma^2)$ to represent the future motion features with the distribution generator $\textsc{DiS}_\theta$, which is supervised by the ground-truth trajectory during training.
The sampled latent variable $z$ from the distribution, together with the planning embedding which carries the current motion features, is decoded with the GRU-based state decoder $\textsc{State}_\theta$ to generate the future motion states $s_{ego}$ across all timesteps.
This resulting ego states $s_{ego}$ are subsequently converted into $K$ coarse trajectory proposals with MLPs.

\begin{gather}
     z \sim p(\mu, \sigma^2); (\mu, \sigma^2) = \textsc{DiS}_{\theta}(H_{p}), z \in \mathbb{R}^{C_z} \\
     s_{ego} = \textsc{StAtE}_{\theta}(z, H_p), s_{ego} \in \mathbb{R}^{F \times 2D}
\end{gather}
where $C_z$ is the latent embedding dimension of the Gaussian distribution.

We further project the future ego motion states to the ego motion queries $Q_{ego}$ in the second decoding stage. 
The trajectory refiner module $\textsc{TrajRefiner}_\theta$ is composed of stacked transformer blocks, where cross-attention is applied to extract future semantics from the predicted visual embeddings.
We transform the high-dimensional future visual embeddings $H_{t+j}$ instead of the direct utilization of predicted latent features $V_{t+j}$ to preserve more reasoning information from the LLM.

\begin{gather}
     Q_{ego} = \textsc{ProJ}_{\theta}(s_{ego}), Q_{ego} \in \mathbb{R}^{F \times C_r}\\
     K_{fut}, V_{fut} = \textsc{ProJ}_{\theta}(H_{t+1:t+F}), K_{fut}, V_{fut} \in \mathbb{R}^{F \times C_r} \\
     s_{ego}^{\ast} = \textsc{TrajRefiner}_\theta(Q_{ego}, K_{fut}, V_{fut}), s_{ego}^{\ast} \in \mathbb{R}^{F \times C_r}
\end{gather}
where $C_r$ is the cross-attention dimension.
The refined motion feature $s_{ego}^{\ast}$ is then decoded into $K$ base trajectories and $K$ offsets through two separate MLP blocks to enhance the generation flexibility and diversity. Finally, the combination of the base trajectory and the offset forms the fine-grained trajectory $\tau_f$.

\subsection{Training Objectives}

LVDrive is trained in an end-to-end manner with a combination of four kinds of loss terms, eliminating the need for a complex multi-stage training process.

For the proposed latent future visual feature prediction task, we employ the weighted sum of the frame-wise cosine similarity loss and the L1 loss. The future visual prediction loss is defined as $\mathcal{L}_{vis} = w_{c}\mathcal{L}_{cosine}+w_{r}\mathcal{L}_{reg}$.
For the two-stage trajectory planning, we adopt the same regression loss $\mathcal{L}_{mse}$, boundary loss $\mathcal{L}_{bd}$ and collision loss $\mathcal{L}_{col}$ for both coarse trajectory proposal and final trajectory, termed as $\mathcal{L}_{plan}$ and $\mathcal{L}_{plan\_r}$. We empirically omit the KL-divergence loss in VAE objectives, as its regularization might overly constrain the multi-modal latent feature space and hinders the learning ability. 

We also follow the previous work to leverage the training loss $\mathcal{L}_{qt}$ of the structured multi-view feature extraction.
And we utilize the cross-entropy loss $\mathcal{L}_{ce}$ for the LLM to generate the fixed special placeholder tokens. 
So that the overall loss is summarized as follows:
\begin{gather}
     \mathcal{L} = \mathcal{L}_{vis} + \mathcal{L}_{plan} + \mathcal{L}_{plan\_r} + \mathcal{L}_{qt} + \mathcal{L}_{ce}.
\end{gather}

\section{Experiments}

\begin{table*}
\centering
\caption{Closed-loop and Open-loop Results of E2E-AD Methods in Bench2Drive under \textbf{base} training set. C/L refers to camera/LiDAR. Avg. L2 is averaged over the predictions in 3 seconds under 2Hz, similar to UniAD. * denote expert feature distillation. IL: Imitation Learning, RL: Reinforcement Learning, DS: Driving Score, SR: Success Rate.}
\vspace{-5pt}
\label{tab: main_res}
\footnotesize
{\setlength{\tabcolsep}{4.7pt}
\begin{tabularx}{\textwidth}{@{} >{\hsize=1.0\hsize\raggedright\arraybackslash}X c c >{\columncolor{gray!10}}c >{\columncolor{gray!10}}c c c c@{}}
\toprule
\multirow{2.2}{*}{Method} & \multirow{2.2}{*}{Scheme} & \multirow{2.2}{*}{Modality} & \multicolumn{4}{c}{Closed-loop Metric} & \color{gray}Open-loop Metric \\
\cmidrule(lr){4-7} \cmidrule(lr){8-8}
& & & DS$\uparrow$ & SR(\%)$\uparrow$ & Efficiency$\uparrow$ & Comfortness$\uparrow$ & \color{gray}Avg. L2 $\downarrow$ \\
\midrule

\multicolumn{8}{c}{\textit{Traditional End-to-End Paradigm}} \\
\midrule

TCP*~\cite{wu2022trajectoryguided} & IL & C & 40.70 & 15.00 & 54.26 & 47.80 & \color{gray}1.70 \\
VAD~\cite{Jiang2023VADVS} & IL & C & 42.35 & 15.00 & 157.94 & 46.01 & \color{gray}0.91 \\
MomAD~\cite{Song2025DontST} & IL & C & 44.54 & 16.71 & 170.21 & 48.63 & \color{gray}0.87 \\
UniAD-Base~\cite{hu2023_uniad} & IL & C & 45.81 & 16.36 & 129.21 & 43.58 & \color{gray}0.73 \\
TCP-traj*~\cite{wu2022trajectoryguided} & IL & C & 59.90 & 30.00 & 76.54 & 18.08 & \color{gray}1.70 \\
DriveDPO~\cite{Shang2025DriveDPOPL} & RL & C & 62.02 & 30.62 & 166.80 & 26.79 & \color{gray}- \\
ThinkTwice*~\cite{jia2023thinktwice} & IL & C & 62.44 & 31.23 & 69.33 & 16.22 & \color{gray}0.95 \\
DriveTransformer-Large~\cite{jia2025drivetransformer} & IL & C & 63.46 & 35.01 & 100.64 & 20.78 & \color{gray}0.62 \\
DriveAdapter*~\cite{Jia2023DriveAdapterBT} & IL & C\&L & 64.22 & 33.08 & 70.22 & 16.01 & \color{gray}1.01 \\
DiffAD~\cite{wang2025diffad} & IL & C & 67.92 & 38.64 & - & - & \color{gray}1.55 \\
Raw2Drive~\cite{yang2026rawdrive} & RL & C & 71.36 & 50.24 & 214.17 & 22.42 & \color{gray}-\\

\midrule

\multicolumn{8}{c}{\textit{Vision-Language-Action Paradigm}} \\
\midrule

Drive$\pi_0$~\cite{Yang2025DriveMoEMF} & IL & C & 60.45 & 30.00 & 168.41 & 14.88 & \color{gray}0.56 \\
ReCogDrive~\cite{li2025recogdrive} & RL & C & 71.36 & 45.45 & 138.18 & 17.45 & \color{gray}- \\
DriveMoE~\cite{Yang2025DriveMoEMF} & IL & C & 74.22 & 48.64 & 175.96 & 15.31 & \color{gray}\textbf{0.38} \\
ORION~\cite{fu2025orion} & IL & C & 77.74 & 54.62 & 151.48 & 17.38 & \color{gray}0.68 \\
MindDrive~\cite{fu2025minddrive} & RL & C & 78.04 & 55.09 & - & - & - \\
UniDrive-WM-AR~\cite{Xiong2026UniDriveWMUU} & IL & C & 79.22 & 56.36 & 158.44 & 28.01 & \color{gray}0.64 \\
UniDrive-WM-AR+Diff~\cite{Xiong2026UniDriveWMUU} & IL & C & 79.31 & 56.42 & 158.65 & 27.93 & \color{gray}0.63 \\

\midrule
\rowcolor[RGB]{230, 242, 255}LVDrive (\textbf{Ours}) & IL & C & \textbf{80.71} & \textbf{58.26} & 155.77 & 14.34 & \color{gray}0.63 \\
\bottomrule
\end{tabularx}}
\vspace{-15pt}
\end{table*}


\subsection{Dataset and Evaluation Metrics}

\paragraph{Dataset.}

We conduct experiments on Bench2Drive dataset~\cite{jia2024bench}, a closed-loop evaluation benchmark built on CARLA v2 simulator~\cite{Dosovitskiy2017CARLAAO} for end-to-end autonomous driving.
For the fair comparison, we adopt the official base training split, which contains 1000 clips uniformly distributed across 44 interactive scenes, 23 weathers and 12 towns.
And we follow the standard split of 950 clips for training and 50 for open-loop validation.
Closed-loop performance is evaluated on the official evaluation protocol, which encompasses 220 short routes.
These routes cover all 44 scenarios with five distinct route instances per scenario under varying weather and town configurations, enabling a comprehensive assessment of a diverse set of challenging driving scenarios and skills.
Additionally, for efficient ablation studies during the model development, we further utilize Dev10~\cite{jia2025drivetransformer}, a compact evaluation subset officially provided, which consists of 10 representative routes.

\paragraph{Evaluation metrics.}
Bench2Drive officially employs five closed-loop evaluation metrics: Driving Score (DS), Success Rate (SR), Efficiency, Comfort and Multi-Ability.
SR quantifies the proportion of routes successfully completed within designated time limit.
DS incorporates both route completion status and violation penalties, where infractions proportionally reduce the score through discount factors.
Efficiency and Comfort respectively assess the speed performance and driving smoothness, and Multi-Ability provides a disentangled assessment of five advanced urban driving skills.
We adopt L2 trajectory error in open-loop evaluation, as used in motion planning~\cite{liao2025diffusiondrive, Cheng2024PLUTOPT, Mei2025HAMFAH}.

\subsection{Implementation Details}

\paragraph{Model Settings.}
\label{implementation}

For the latent future prediction, we set N as 256, indicating that each future feature frame is represented by 256 tokens.
Ground-truth latent visual features in 6 future timesteps are provided by the pre-trained VQGAN-ImageNet~\cite{esser2020taming} with the codebooksize as 16384. We utilize the quantized code embeddings instead of the discrete code indices as the visual supervision signal.
As for the planning task, 6 multi-modal trajectories are generated corresponding to the 6 navigation commands defined in Bench2Drive.
Additional implementation details are provided in the Appendix.

\paragraph{Training and Evaluation Process.}

Our experiments are conducted on 32 NVIDIA H20 GPUs with 96 GB of memory.
We employ EVA-02-L~\cite{Fang2023EVA02AV} and QT-Former~\cite{fu2025orion} as the vision encoder. Vicuna v1.5~\cite{Zheng2023JudgingLW} is utilized as the core LLM and fine-tuned with LoRA~\cite{Hu2021LoRALA}. 
We adopt the pre-trained weights in the first stage of ORION~\cite{fu2025orion}, which is trained on VQA pairs from Chat-B2D to align the vision space with the reasoning space.
Our LVdrive is trained in an end-to-end manner for 6 epochs, eliminating the complex multi-stage training process.
In the evaluation, the fixed visual placeholder and planning placeholder token sequence is pre-filled, so that the predicted future scene and action are obtained in a single forward process with parallel decoding instead of the autoregressive decoding, improving the inference efficiency significantly.

\subsection{Main Results}

Table~\ref{tab: main_res} summarizes the closed-loop evaluation results on the Bench2Drive benchmark. 
Our LVDrive consistently outperforms all existing approaches with 80.71 Driving Score and 58.26\% Success Rate, including both conventional end-to-end planners and VLA-based methods, under both imitation learning and reinforcement learning paradigms.
Notably, a particularly instructive comparison is with UniDrive-WM~\cite{Xiong2026UniDriveWMUU}, which employs autoregressive generation of texture-level future images as an auxiliary task and relies on a complex multi-stage training pipeline. 
The substantial improvement of LVDrive over UniDrive-WM highlights the advantage of our latent future representation learning.
By learning in a latent space rather than pursuing high-fidelity pixel reconstruction, our method captures semantically richer scene features that are more directly beneficial for motion planning.

\begin{table*}[t] 
\centering
\caption{Multi-Ability Results of E2E-AD Methods under \textbf{base} training set. * denote expert feature distillation. C/L refers to camera/LiDAR. IL: Imitation Learning, RL: Reinforcement Learning.}
\label{tab: multi_ab_res}
\footnotesize
{\setlength{\tabcolsep}{2.2pt}
\begin{tabularx}{\textwidth}{@{} >{\hsize=1.0\hsize\raggedright\arraybackslash}X c c c c c c c >{\columncolor{gray!10}}c @{}}
\toprule
\multirow{2.3}{*}{Method} & \multirow{2.3}{*}{Scheme} & \multirow{2.3}{*}{Modality} & \multicolumn{6}{c}{Ability (\%) $\uparrow$} \\
\cmidrule(lr){4-9}
& & & Merging & Overtaking & Emergency Brake & Give Way & Traffic Sign & Mean \\
\midrule

\multicolumn{8}{c}{\textit{Traditional End-to-End Paradigm}} \\
\midrule

TCP*~\cite{wu2022trajectoryguided} & IL & C & 16.18 & 20.00 & 20.00 & 10.00 & 6.99 & 14.63 \\
UniAD-Base~\cite{hu2023_uniad} & IL & C & 14.10 & 17.78 & 21.67 & 10.00 & 14.21 & 15.55 \\
VAD~\cite{Jiang2023VADVS} & IL & C & 8.11 & 24.44 & 18.64 & 20.00 & 19.15 & 18.07 \\
TCP-traj*~\cite{wu2022trajectoryguided} & IL & C & 8.89 & 24.29 & 51.67 & \underline{40.00} & 46.28 & 34.22 \\
ThinkTwice*~\cite{jia2023thinktwice} & IL & C & 27.38 & 18.42 & 35.82 & \textbf{50.00} & 54.23 & 37.17 \\
DriveTransformer-Large~\cite{jia2025drivetransformer} & IL & C & 17.57 & 35.00 & 48.36 & \underline{40.00} & 52.10 & 38.60 \\
DiffAD~\cite{wang2025diffad} & IL & C & 30.00 & 35.55 & 46.66 & \underline{40.00} & 46.32 & 38.79 \\
DriveAdapter*~\cite{Jia2023DriveAdapterBT} & IL & C\&L & 28.82 & 26.38 & 48.76 & \textbf{50.00} & 56.43 & 42.08 \\
Raw2Drive~\cite{yang2026rawdrive} & RL & C & \textbf{43.35} & 51.11 & 60.00 & \textbf{50.00} & 62.26 & 53.34 \\

\midrule

\multicolumn{8}{c}{\textit{Vision-Language-Action Paradigm}} \\
\midrule

Drive$\pi_0$~\cite{Yang2025DriveMoEMF} & IL & C & 29.35 & 36.58 & 48.83 & \underline{40.00} & 54.45 & 41.84 \\
DriveMoE~\cite{Yang2025DriveMoEMF} & IL & C & 34.67 & 40.00 & 65.45 & \underline{40.00} & 59.44 & 47.91 \\
ReCogDrive~\cite{li2025recogdrive} & RL & C & 29.73 & 20.00 & 69.09 & 20.00 & 71.34 & 42.03 \\
ORION~\cite{fu2025orion} & IL & C & 25.00 & 71.11 & 78.33 & 30.00 & 69.15 & 54.72 \\
MindDrive~\cite{fu2025minddrive} & RL & C & 32.89 & \textbf{75.56} & 68.33 & \textbf{50.00} & 57.89 & 56.94 \\
UniDrive-WM-AR~\cite{Xiong2026UniDriveWMUU} & IL & C & 28.81 & 74.04 & \underline{79.84} & \underline{40.00} & 71.30 & 59.00 \\
UniDrive-WM-AR+Diff~\cite{Xiong2026UniDriveWMUU} & IL & C & 29.97 & \underline{74.64} & \textbf{79.98} & \underline{40.00} & \underline{71.54} & 59.23 \\
\midrule
\rowcolor[RGB]{230, 242, 255}%
LVDrive (\textbf{Ours}) & IL & C & \underline{39.74} & 68.89 & 71.67 & 20.00 & \textbf{74.21} & 54.90 \\
\bottomrule
\end{tabularx}}
\vspace{-15pt}
\end{table*}

We further report the Multi-Ability evaluation results in Table~\ref{tab: multi_ab_res}. 
LVDrive achieves the highest Traffic Sign score of 74.21\% and the second-best Merging score of 39.74\%, together with competitive performance on Overtaking and Emergency Brake. 
These results demonstrate the strong scene understanding and reasoning capability enabled by latent scene representation learning. 
Since most evaluation scenarios correspond to Traffic Sign and Merging, the gains in these two skills largely account for the superior DS and SR.
For the Give Way skill, where the ego vehicle is typically required to yield to an emergency vehicle approaching from behind, LVDrive often fails to complete the maneuver.
This failure is likely due to the incorporation of the front-view future scene prediction task, which leads the model's attention predominantly toward front-view visual cues and thereby limits its capacity to capture interactions occurring behind.
More detailed analysis with qualitative results is provided in the Appendix~\ref{a_vis}.

\subsection{Ablation Study}

For ablation studies, we utilize \textit{Dev10}, a dedicated subset officially provided and recommended by Bench2Drive benchmark. 
It consists of 10 clips, each selected from a distinct representative scenario.
The small-scale subset offers a diverse yet low-variance evaluation signal that prevents overfitting the full 220 testing routes during the model development.

\begin{table}[t]
    \centering
	\begin{minipage}{0.45\textwidth}
		\centering
            \setlength{\abovecaptionskip}{0pt}            
\caption{Ablation study on core components of LVDrive on Dev10 set. 
"Latent Vis." indicates the latent future frame prediction task.
"One-stage Dec." and "Two-stage Dec." indicate the trajectory decoding strategy. 
DS: Driving Score; SR: Success Rate.
}
\vspace{0.05em}
\label{tab: ab_key}
\footnotesize
\setlength{\tabcolsep}{0.4pt}
\begin{tabular}{c c c c c c}
\toprule[1.5pt]
\multirow{2.6}{*}{\shortstack{Model\\Variants}} & 
\multirow{2.6}{*}{\shortstack{Latent\\Vis.}} &
\multirow{2.6}{*}{\shortstack{One-stage\\Dec.}} &
\multirow{2.6}{*}{\shortstack{Two-stage\\Dec.}} &
\multicolumn{2}{c}{Closed-loop} \\
\cmidrule(lr){5-6}
& & & & DS $\uparrow$ & SR $\uparrow$ \\
\midrule
$\mathcal{M}_{base}$ &            &     &       & 65.25 & 4/10 \\
$\mathcal{M}_{vis}$ & \checkmark &      &      & 66.31 & 3/10 \\
$\mathcal{M}_{one}$ & \checkmark & \checkmark       &     & 60.43 & 3/10 \\
\rowcolor[RGB]{230,230,230} 
\textbf{LVDrive (ours)} & \checkmark & & \checkmark & \textbf{82.39} & \textbf{7/10} \\
\bottomrule[1.5pt]
\end{tabular}%
        \end{minipage}
        \hfill
        \begin{minipage}{0.53\textwidth}
    	\centering
            \setlength{\abovecaptionskip}{0pt}
\caption{Ablation study on different latent visual feature supervisions.
"Vision Supervision" indicates the utilized pre-trained visual backbone.
"Feature Dim." indicates the dimension of the ground-truth latent feature. 
DS: Driving Score; SR: Success Rate.
}
 \vspace{0.15em}
\label{tab: ab_vis}
\footnotesize
\setlength{\tabcolsep}{0.4pt}
\begin{tabular}{c c c c c}
\toprule[1.5pt]
\multirow{2}{*}{Model Variants} & 
\multirow{2}{*}{Vision Supervision} &
\multirow{2}{*}{Feature Dim.} &
\multicolumn{2}{c}{Closed-loop} \\
\cmidrule(lr){4-5}
& & & DS $\uparrow$ & SR $\uparrow$ \\
\midrule
$\mathcal{M}_{base}$ &    -        &   -         & 65.25 & 4/10 \\
$\mathcal{M}_{1}$ &     Internal Vision Enc.       &     1024       & 65.42 & 4/10 \\
$\mathcal{M}_{2}$ &   MoVQGAN~\cite{Wang2024Emu3NP}         &     4       & 59.91 & 3/10 \\
$\mathcal{M}_{3}$ & DINOv3-Large~\cite{Simeoni2025DINOv3} &    1024        & 71.72 & 5/10 \\
\rowcolor[RGB]{230,230,230} 
\textbf{LVDrive (ours)} & VQGAN-ImageNet~\cite{esser2020taming} & 256 & \textbf{82.39} & \textbf{7/10} \\
\bottomrule[1.5pt]
\end{tabular}%
	\end{minipage}

\vspace{-16.5pt}
\end{table}

\paragraph{Effects of key components.}

Table~\ref{tab: ab_key} reports the effectiveness of two key designs in LVDrive.
The baseline model $\mathcal{M}_{base}$, presented in the first row, is trained to plan solely from action labels, analogous to the variants used in the second training phase of~\cite{fu2025orion}.
We then directly incorporate the latent future scene prediction task as an auxiliary objective, denoted $\mathcal{M}_{vis}$. 
While the Driving Score improves, a decline in Success Rate is observed under this multi-task setting. 
We speculate that this degradation arises because motion planning and future scene prediction operate in distinct action and vision feature spaces that are not inherently aligned. Consequently, naively combining them introduces conflicting optimization signals.
To address this misalignment, we first investigate a one-stage decoding variant $\mathcal{M}_{one}$. 
The planning embedding $H_p$ is directly used as the ego motion query to capture future features through the cross-attention mechanism and then fed into the VAE-base planner, constraining both tasks to optimize toward a shared objective. However, this direct integration negatively impacts motion feature learning, resulting in further performance decline.
We therefore adopt a two-stage trajectory decoding strategy. This design first leverages action labels to stabilize the learned action features for decoding coarse proposals, and subsequently aggregates the learned future scene features to refine the final trajectory.
With this design, LVDrive ultimately achieves significant and consistent performance gains.

\paragraph{Effects of various latent visual supervision signals.}

The vision backbone responsible for supplying future scene supervision signals plays a crucial role in our framework.
We conduct experiments with several other alternative backbones, as summarized in Table~\ref{tab: ab_vis}.
As shown in $\mathcal{M}_1$, we first experiment with the internal vision encoder that processes multi-view image input.
This yields a slight improvement in Driving Score but no gain in Success Rate.
We attribute this to the fact that the encoder is jointly trained with multiple objectives during the training, limiting its visual representation capacity compared to the external large-scale pretrained backbones.

We then explore the pre-trained backbones.
One candidate is the post-quantized code embedding from the MoVQGAN vision encoder of Emu3~\cite{Wang2024Emu3NP}, the aim of which is to tokenize images into a discrete space.
So the code embedding preserves the critical and highly compressed features to recover the original image.
However, this variant leads to a performance decline shown in $\mathcal{M}_2$, likely due to the low feature dimension of the ground-truth future frame compared to the LLM's hidden state, making it difficult to transfer sufficient information.
In contrast, the features from DINOv3~\cite{Simeoni2025DINOv3} bring clear gains in $\mathcal{M}_3$, confirming the effectiveness of latent future scene representation learning.
Nevertheless, compared to VQGAN-ImageNet, the feature of DINOv3-Large is excessively rich, potentially introducing redundant information as noise into the shared vision-action feature space.

\paragraph{Effectiveness of two-stage trajectory decoding.}

We further validate the effectiveness of the proposed two-stage trajectory decoding strategy in Table~\ref{tab: ab_dec}.
Since both the coarse trajectory proposal and the final fine-grained trajectory are generated during inference, we evaluate the closed-loop performance of each to examine how future scene features are progressively aggregated through the two stages.
The coarse trajectory in $\mathcal{M}_{coarse}$, which is directly decoded from the planning token, already achieves a substantial improvement over the baseline $\mathcal{M}_{base}$.
This confirms that the planning embedding initially captures effective future scene representations through future-aware reasoning.
The performance is further boosted when the motion feature is explicitly refined with latent visual embeddings in the second stage, leading to the final trajectory.

\paragraph{Efficiency of single forwarding process.}

In contrast to autoregressive paradigms that generate visual and action tokens sequentially, LVDrive processes pre-filled placeholder tokens for latent visual features and actions in a single forward step, which in principle leads to higher inference efficiency. 
We report the measured inference speed in Table~\ref{tab: ab_time}. 
Measurements are performed on the official Bench2Drive validation set using the NVIDIA H20 GPU with a batch size of one.
LVDrive achieves an inference speed approximately twice that of $\mathcal{M}_{base}$, despite the overhead of the latent future scene prediction task.
In comparison, a hypothetical autoregressive baseline $\mathcal{M}_{AR}$, which is configured to generate vision and action tokens of equivalent length, results in an order of magnitude slower, incurring prohibitive computational costs for the autonomous driving deployment.

\subsection{Qualitative Results}

\begin{figure*}[t]
    \centering
    \includegraphics[width=0.99\textwidth, trim=0 200 40 0]{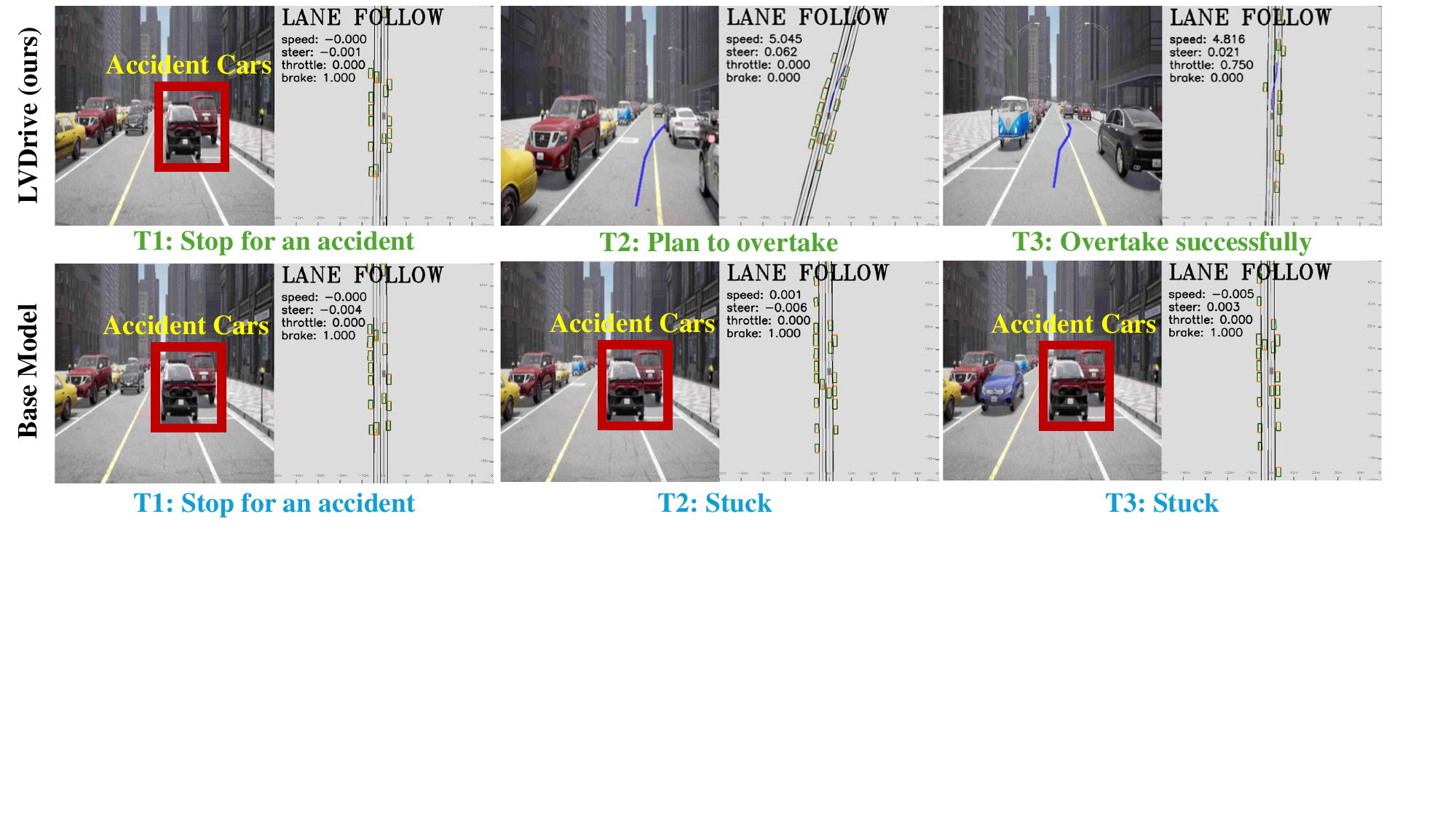}
    \caption{
    \textbf{Qualitative results of our LVDrive and $\mathcal{M}_{base}$ in an Overtaking scenario from Bench2Drive.} 
    The ego vehicle encounters an accident ahead that blocks its driving lane, while a steady stream of oncoming traffic occupies the adjacent lane.
    The $\color{blue}{blue}$ line denotes the generated trajectory.
    The ego vehicle controlled by $\mathcal{M}_{base}$ becomes immobilized at the accident site.
    In contrast, our LVDrive successfully and smoothly overtakes the accident cars with the safe planning.
    }
    \label{fig:vis}
    \vspace{-13pt}
\end{figure*}

The qualitative results of LVDrive and $\mathcal{M}_{base}$ are presented in Figure~\ref{fig:vis}.
In this scene, the ego vehicle is tasked with navigating a challenging scenario, which demands highly precise planning.
The ego vehicle encounters the accident ahead, with oncoming traffic continuously approaching from the opposite direction in the adjacent lane.
$\mathcal{M}_{base}$ performs unsatisfactorily in this scenario, where the ego vehicle halts at the accident site and remains stranded there, unable to plan a drivable path.
In contrast, enhanced by its future scene prediction capability, LVDrive successfully maneuvers past the accident cars and navigates the lane safely, undisturbed by the oncoming traffic.
This comparison demonstrates that LVDrive possesses stronger scene understanding and reasoning abilities, yielding more robust and reliable planning performance.
Additional results are provided in the Appendix~\ref{a_vis}.

\begin{table}[t]
    \centering
	\begin{minipage}{0.6\textwidth}
		\centering
            \setlength{\abovecaptionskip}{0pt}
\caption{Ablation study on two-stage decoding strategy.
"Trajectory Type" indicates whether coarse proposal or fine-gained trajectory is utilized. 
}
\label{tab: ab_dec}
\footnotesize
\setlength{\tabcolsep}{9.5pt}
\begin{tabular}{c c c c}
\toprule[1.3pt]
\multirow{2.6}{*}{Model Variants} & 
\multirow{2.6}{*}{Trajectory Type} &
\multicolumn{2}{c}{Closed-loop} \\
\cmidrule(lr){3-4}
& & DS $\uparrow$ & SR $\uparrow$ \\

\midrule
$\mathcal{M}_{base}$ &    -    & 65.25 & 4/10 \\
$\mathcal{M}_{coarse}$ &    coarse proposal               & 73.22 & 5/10 \\
\rowcolor[RGB]{230,230,230} 
\textbf{LVDrive (ours)} & fine-grained trajectory & \textbf{82.39} & \textbf{7/10} \\

\bottomrule[1.3pt]
\end{tabular}%
        \end{minipage}
        \hfill
        \begin{minipage}{0.38\textwidth}
    	\centering
            \setlength{\abovecaptionskip}{0pt}
\caption{Comparison of inference speed with different model variants. The inference time is reported in seconds. 
}
\vspace{0.2em}
\label{tab: ab_time}
\footnotesize
\setlength{\tabcolsep}{10pt}
\begin{tabular}{c c c}
\toprule[1.1pt]
Model Variants & Inference Time  \\
\midrule
$\mathcal{M}_{base}$ &    0.93s      \\
$\mathcal{M}_{AR}$ &    36.62s     \\
\rowcolor[RGB]{230,230,230} 
\textbf{LVDrive (ours)} & 2.03s   \\

\bottomrule[1.1pt]
\end{tabular}%
	\end{minipage}
\vspace{-16pt}
\end{table}

\section{Conclusion}

In this work, we introduce \textbf{LVDrive}, a VLA framework enhanced with latent future visual representation learning, guided by auxiliary semantic supervision from a pretrained VQGAN model.
LVDrive performs future-aware reasoning that jointly predicts future scene features and motion features in a single forward pass, efficiently using pre-filled placeholder tokens.
The trajectory is generated through the dedicated two-stage decoding strategy, which is explicitly refined with the learned future visual representations.
Extensive experiments on the challenging Bench2Drive benchmark demonstrate that LVDrive outperforms existing VLA frameworks, including those relying on sparse action supervision and those incorporating image-reconstruction-based world modeling.
\paragraph{Limitations.}
One limitation of the current framework is that it primarily leverages the reasoning capability of the underlying language space, without explicitly investigating how language supervision could further enhance visual and action representation learning.
Another limitation concerns the choice of the pre-trained vision backbone. 
Although LVDrive already achieves strong performance with general-purpose backbones, we believe that adopting vision foundation models specifically pre-trained on large-scale autonomous driving data could further boost planning performance.

\section{Acknowledgment}
\paragraph{Data usage statement.} 
This paper uses the external dataset resources. We utilize the dataset provided in the Bench2Drive project (source: \hyperlink{}{\textcolor{blue}{https://github.com/Thinklab-SJTU/Bench2Drive}}), which is licensed under the \textbf{CC BY-NC-ND 4.0}.
The authors confirm that the usage of the above data in this paper is strictly limited to academic research and has not involved any commercial activities.

\bibliographystyle{ieeenat_fullname}
\bibliography{main}

\newpage
\appendix

\section{Technical appendices and supplementary material}
We first provide more detailed implementations of our LVDrive in the following Section~\ref{a1} to Section~\ref{a3}.
Then we provide more qualitative results and comparisons in Section~\ref{a_vis}. 

\subsection{Latent Vision Decoding}
\label{a1}

\paragraph{Vision decoder $\textsc{ViS}_{\theta}$.}

The visual decoder $\textsc{ViS}_{\theta}$ extracts latent semantic features $V_{t+j} \in \mathbb{R}^{N \times C_v}$ from the vision embedding $H_{t+j} \in \mathbb{R}^{N \times D}$ produced by LLM.
Here, $D$ is set to 4096, matching the hidden dimension of the LLM, and $C_v$ is set to 256, corresponding to the post-quantized embedding dimension of VQGAN-ImageNet~\cite{esser2020taming}. 
The lightweight vision decoder $\textsc{ViS}_{\theta}$ consists of a single linear layer that projects from 4096 to 256 dimensions, with no activation function applied.

\paragraph{Ground-truth future scene feature generation.}
To obtain the ground-truth supervision targets, we feed the future front-view images with the width and height of $256 \times 256$ at the next six timesteps into VQGAN-ImageNet~\cite{esser2020taming}, which produces latent visual features of shape $16 \times 16 \times 256$.
Layer normalization is then applied to these ground-truth features to stabilize training.

\subsection{Two-stage Trajectory Decoding}
\label{a2}

\paragraph{VAE-based planner.}

As described in Section~\ref{two-stage decoding}, the planning embedding is first mapped to a Gaussian distribution by the distribution generator $\textsc{DiS}_\theta$.
This generator employs three $1\times1$ 1D convolutional layers with ReLU activations~\cite{Agarap2018DeepLU}, which compress the 4096-dimensional planning embedding to 2048 dimensions.
An average pooling layer followed by an additional $1\times1$ convolutional layer then produces the mean and log standard deviation of the distribution, each with a latent dimension $C_z$ as 32.
During training, the ground-truth ego trajectory is concatenated with the planning embedding as the input to generate the distribution to stabilize distribution learning.
A latent code of dimension 32 is then sampled and fed into the GRU-based state decoder $\textsc{State}_\theta$, which generates the future motion states.

The sampled latent code is first repeated six times to align with the six prediction timesteps.
The state decoder $\textsc{State}_\theta$ consists of four GRU layers with a latent dimension of 1024, followed by three MLP layers with ReLU activations with the output dimension of 4096. 
To preserve the current motion information, the planning embedding is also repeated six times and concatenated with the output future motion states produced by the state decoder.
The resulting ego state $s_{ego}$ thus has a shape of 8192.

This state is subsequently passed through three additional MLP layers with ReLU activations to generate the multi-modal coarse trajectory.

\paragraph{Trajectory refiner.}

The core of the trajectory refinement module is a two-layer transformer block with a cross-attention mechanism, operating with a latent dimension $C_r$ as 1024.
The ego state serves as the learnable query, and the future visual embedding acts as the key and value, both of which are first projected to match the latent dimension.
The projector $\textsc{ProJ}_{\theta}$ consists of a single linear layer, followed by layer normalization and GELU activation~\cite{Hendrycks2016GaussianEL}.
The updated ego motion state $s_{ego}^{\ast}$ is then fed into two separate MLP blocks with identical architectures, which generate the base trajectory and the offset to enhance generation diversity.
Each MLP block contains two linear layers with layer normalization and GELU activation.
The sum of the base trajectory and the offset yields the final trajectory.

\subsection{Vision Encoding with the Multi-View Image Input}
\label{a3}

We adopt the same multi-view visual feature extraction and compression procedure as~\cite{fu2025orion}, to which we refer readers for more detailed explanations.

In brief, image tokens are first extracted by the vision encoder EVA-02-L~\cite{Fang2023EVA02AV}.
The QT-Former module then employs three types of learnable queries, termed as perception queries, scene queries, and history queries, to capture structured visual features, compressed current scene features, and historical scene context, respectively.
The module which performs critical object detection, traffic state prediction, and motion prediction of surrounding agents, is supervised by ground-truth labels during the training, yielding the auxiliary loss $\mathcal{L}_{qt}$.
The compressed current and historical scene features, rather than the heavy image tokens, are subsequently fed into the core LLM.

\subsection{Additional Qualitative Results}
\label{a_vis}

We provide more qualitative comparisons of our LVDrive and the baseline model $\mathcal{M}_{base}$. 
We visualize the multi-view image (left) and the structured BEV view (right) in each scenario. 
The {\color{blue}blue} line denotes the generated trajectory. 
In the BEV view, the ego vehicle is shown as the gray box. 
The detected surrounding vehicles are shown as green boxes with an orange heading indicator that marks their moving direction.

\paragraph{Multi-ability evaluation.}
Bench2Drive defines five advanced driving skills for evaluation: Emergency Brake, Merging, Traffic Sign, Overtaking, and Give Way. 
Each evaluation scenarios is mapped to one or more of these skills, and the distribution across skills is not uniform, as shown in Table~\ref{tab: appen1}.
\begin{table*}[h]
\centering
\setlength{\abovecaptionskip}{0pt}
\caption{Evaluation scenario distribution for five driving skills. 
}
\label{tab: appen1}
\footnotesize
\setlength{\tabcolsep}{9.5pt}
\begin{tabular}{c c c c c c}
\toprule
 & Traffic Sign & Merging & Emergency Brake & Overtaking & Give Way \\
\midrule
Numbers of Scenarios & 18  & 16 & 12 & 9 & 2   \\

\bottomrule
\end{tabular}%
\end{table*}

As shown in Table~\ref{tab: multi_ab_res}, LVDrive achieves strong performance on four of the five driving skills, with the exception of Give Way.
We present qualitative results for the Emergency Brake, Merging, Traffic Sign, and Overtaking scenarios to illustrate this effectiveness.
Additionally, we provide an analysis of a representative failure case in the Give Way scenario to understand the limitations of our approach.

\begin{figure*}[ht]
    \centering
    \includegraphics[width=0.95\textwidth, trim=0 30 0 0]{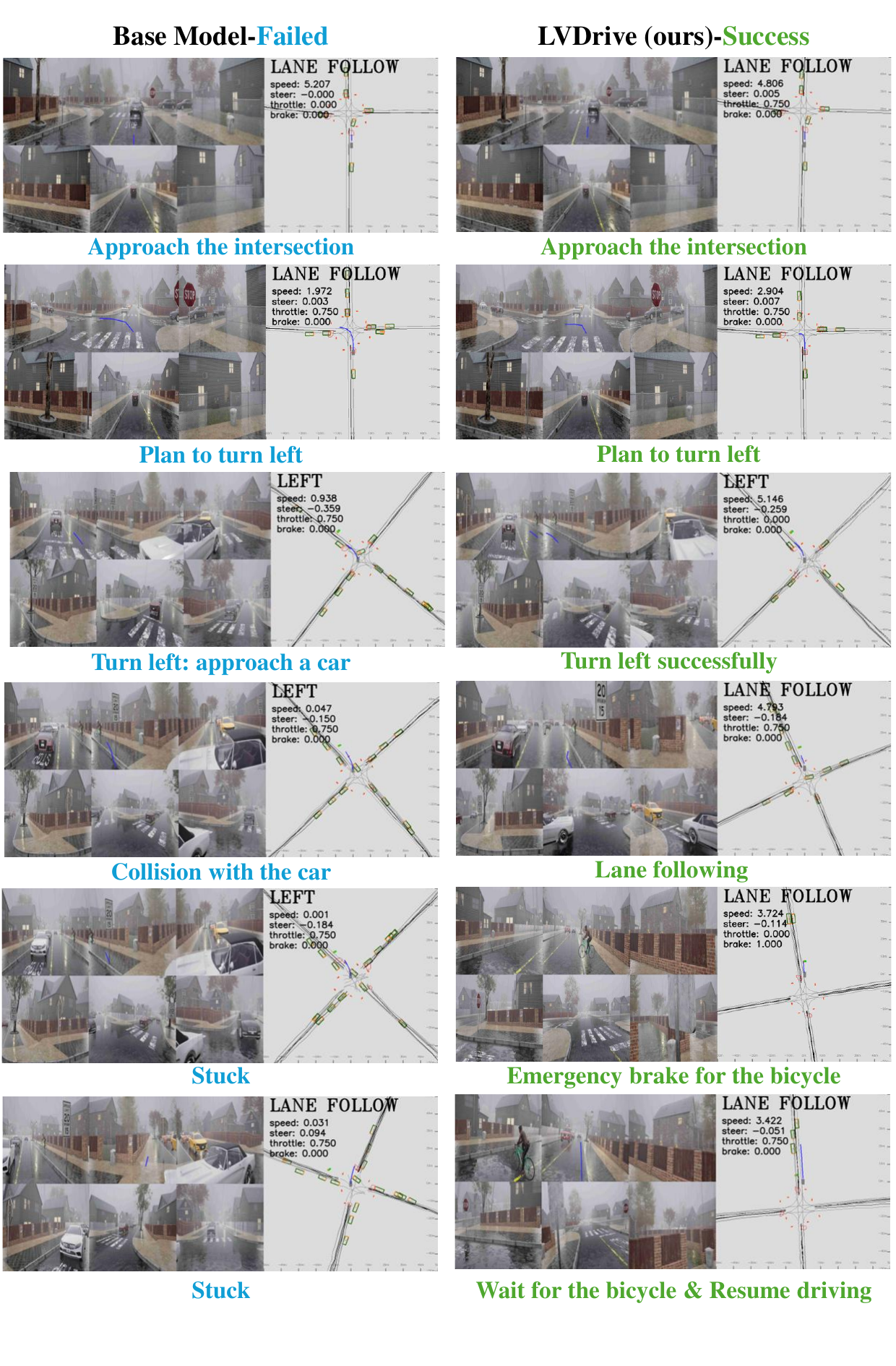}
    \caption{
    \textbf{Qualitative results of our LVDrive and $\mathcal{M}_{base}$ in an Emergency Brake scenario.} 
    The ego vehicle is required to first perform an unprotected left turn at an intersection without a traffic light, then yield to a bicycle crossing its path.
    Upon encountering the crossing bicycle, the ego vehicle should execute an emergency brake, wait for the bicycle to clear the road, and subsequently resume driving.
    The ego vehicle of $M_{base}$ fails to turn, resulting in the collision with the car at the intersection.
In contrast, LVDrive successfully navigates the entire sequence, with completing the turn, braking for the bicycle, and safely resuming driving, which demonstrates robust planning performance in this challenging scenario.
    }
    \label{fig:vis_a3}
\end{figure*}

\paragraph{Performance on Emergency Brake.}
The qualitative results and comparisons in an \textit{Emergency Brake} scenario are shown in Figure~\ref{fig:vis_a3}.
This scenario presents a particularly challenging task that demands both precise spatial reasoning and timely planning.
The ego vehicle is first required to execute an unprotected left turn at an intersection without a traffic light, which also requires careful judgment of gaps in the oncoming traffic.
Immediately after completing the turn, the ego vehicle encounters a bicycle that unexpectedly crosses the path, forcing an abrupt transition from turning to braking.
The ego vehicle must execute an emergency brake with minimal hesitation, wait for the bicycle to fully clear the road, and subsequently resume the driving.
The ego vehicle controlled by $\mathcal{M}_{base}$ fails at the very first step. It cannot complete the unprotected left turn and collides with another vehicle at the intersection.
In contrast, LVDrive successfully navigates the entire sequence by completing the unprotected turn, braking promptly for the crossing bicycle and safely resuming driving.
\textbf{The performance demonstrates how the future scene prediction capability enables LVDrive to anticipate sudden events and plan accordingly.}

\begin{figure*}[ht]
    \centering
    \includegraphics[width=0.95\textwidth, trim=0 30 0 0]{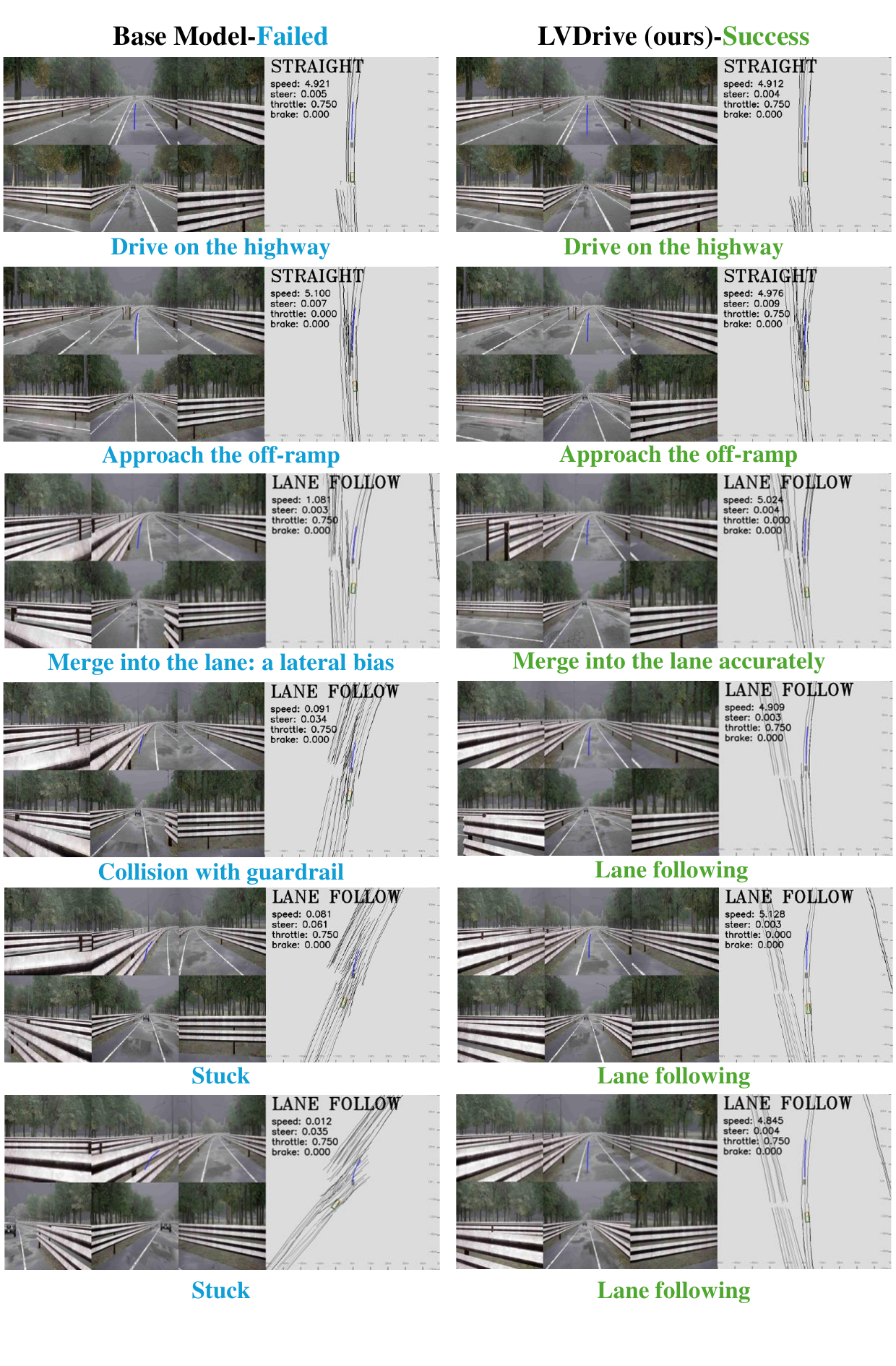}
    \caption{
    \textbf{Qualitative results of our LVDrive and $\mathcal{M}_{base}$ in a Merging scenario.} 
    Navigating highway exits and merging onto narrow roads requires precise perception of the road layout and fine-grained trajectory planning.
The ego vehicle of $\mathcal{M}_{base}$ fails to capture the exact road boundary, leading to a lateral deviation that results in a collision with the guardrail.
In contrast, LVDrive accurately perceives the off-ramp geometry and plans a safe trajectory to merge into the narrow road without incident.
    }
    \label{fig:vis_a2}
\end{figure*}

\paragraph{Performance on Merging.}
The qualitative results and comparisons in a \textit{Merging} scenario are shown in Figure~\ref{fig:vis_a2}.
Navigating highway exits and merging onto narrow roads presents a demanding planning challenge, where the ego vehicle must simultaneously perceive the off-ramp geometry, identify the merging point, and execute a precise lateral maneuver into a confined space.
The ego vehicle controlled by $\mathcal{M}_{base}$ fails to capture the exact road boundary and generates a trajectory with a lateral bias that ultimately leads to a collision with the guardrail.
This failure suggests that without a powerful scene understanding ability, the baseline model struggles to reason about the spatial constraints imposed by the complex road layout.
In contrast, LVDrive accurately perceives the available merging space, plans a safe and smooth trajectory, and successfully merges into the narrow road without incident.
\textbf{This result demonstrates that latent visual representation learning provides a stronger spatial understanding capability, which is critical for precise maneuvers in spatially constrained scenarios.}

\begin{figure*}[ht]
    \centering
    \includegraphics[width=0.95\textwidth, trim=0 30 0 0]{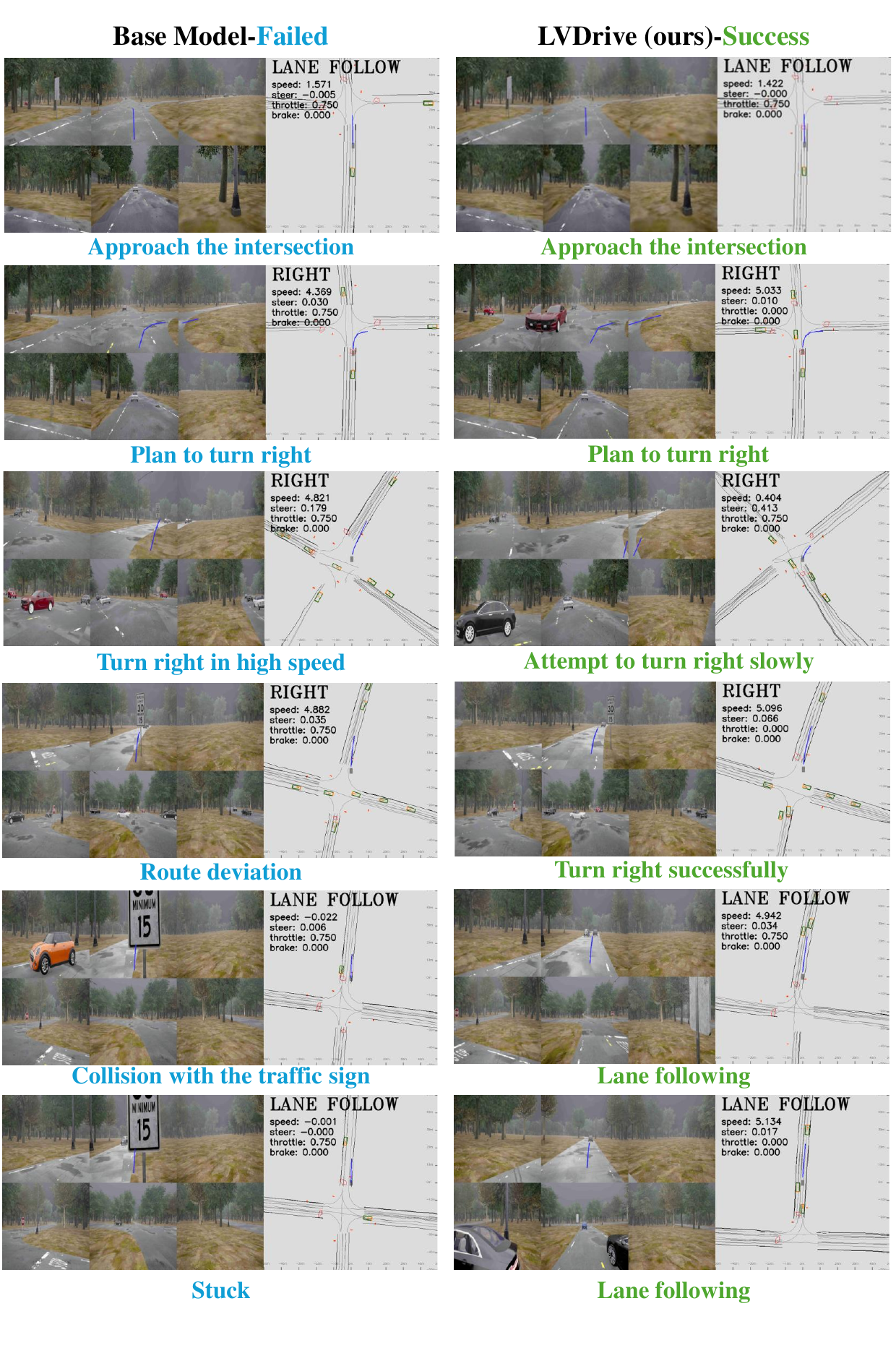}
    \caption{
    \textbf{Qualitative results of our LVDrive and $\mathcal{M}_{base}$ in a Traffic Sign scenario.}
    The ego vehicle is required to make a right turn at a non-signalized intersection without traffic lights or signs, while avoiding collisions with surrounding vehicles.
The ego vehicle controlled by $\mathcal{M}_{base}$ plans an inaccurate trajectory that deviates from the intended route and ultimately collides with a traffic sign.
In contrast, LVDrive successfully and carefully plans a safe trajectory, navigating through the junction without incident.
    }
    \label{fig:vis_a4}
\end{figure*}

\paragraph{Performance on Traffic Sign.}
The qualitative results and comparisons in a \textit{Traffic Sign} scenario are shown in Figure~\ref{fig:vis_a4}.
The ego vehicle is required to make a right turn at a non-signalized intersection without traffic lights or signs, and simultaneously avoid collisions with surrounding vehicles navigating through the junction.
This scenario poses a dual challenge: the model must accurately perceive the unstructured road geometry, including the intersection layout and traffic sign placement, and track the behavior of nearby vehicles to identify safe turning gaps.
The ego vehicle controlled by $\mathcal{M}_{base}$ plans an inaccurate trajectory that deviates from the intended route and ultimately collides with a traffic sign, suggesting a failure to correctly interpret the spatial layout of the intersection.
In contrast, LVDrive successfully captures both the precise road topology and the motion of surrounding dynamic cars, and plans a carefully executed trajectory that navigates through the junction without incident.
\textbf{This highlights LVDrive's enhanced spatial understanding and reasoning of road structures and dynamic agents in the highly-interactive scenario.}

\begin{figure*}[ht]
    \centering
    \includegraphics[width=0.95\textwidth, trim=0 30 0 0]{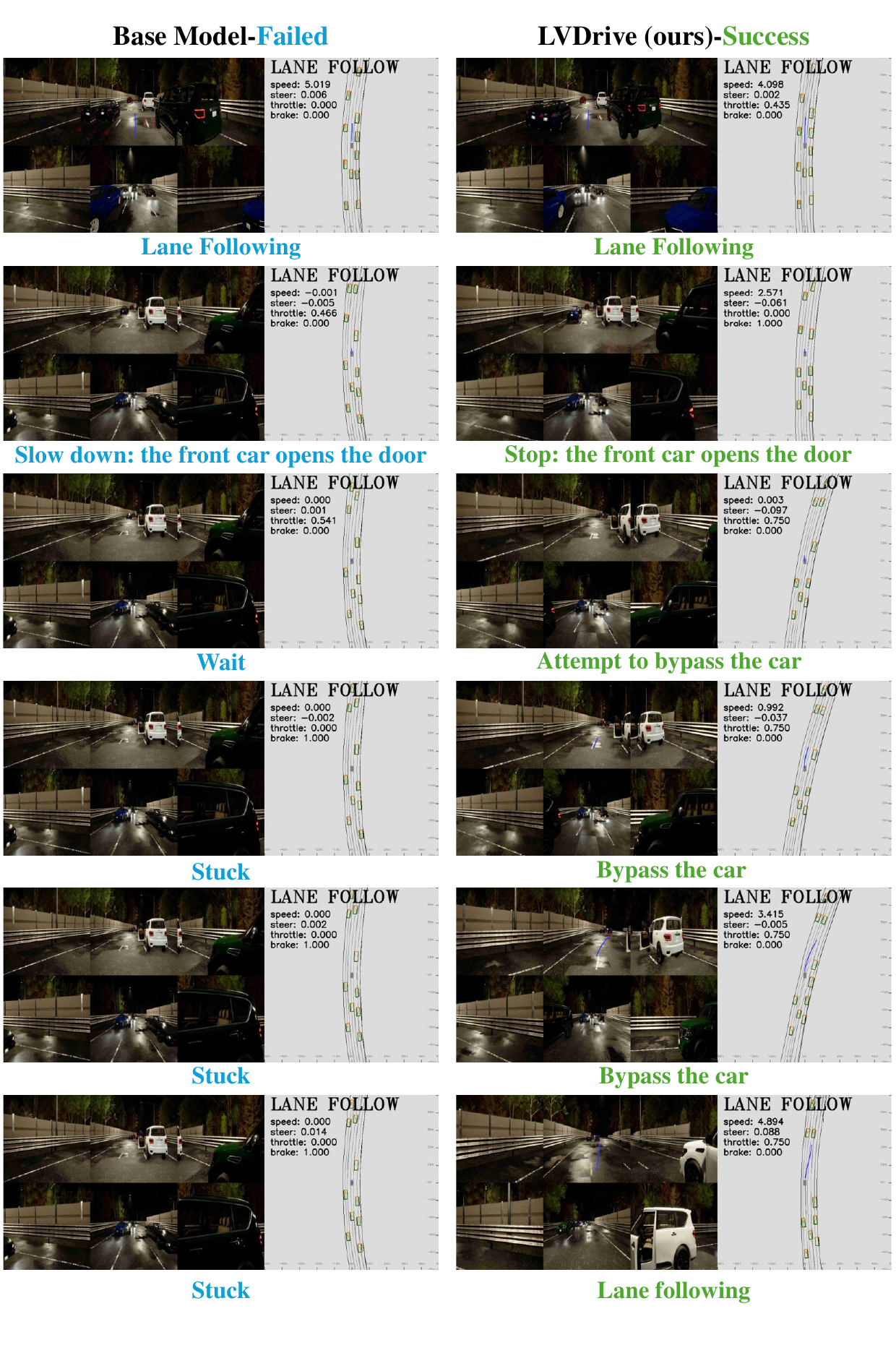}
    \caption{
    \textbf{Qualitative results of our LVDrive and $\mathcal{M}_{base}$ in an Overtaking scenario.}
    The ego vehicle drives forward in its lane, and the front car in the adjacent lane stops and opens the door, which blocks the driving lane.
    The ego vehicle of $\mathcal{M}_base$ stops and gets stuck in the place. 
    In contrast, our LVDrive successfully plans the safe trajectory to bypass the front car with the open door and drives forward continuously.}
    \label{fig:vis_a1}
\end{figure*}

\paragraph{Performance on Overtaking.}
The qualitative results and comparisons in another \textit{Overtaking} scenario are shown in Figure~\ref{fig:vis_a1}.
This scenario presents a subtle yet demanding perception challenge. While the ego vehicle is driving forward, a car parked in the adjacent lane suddenly opens its door, significantly deforming its original contour and partially intruding into the ego vehicle's driving path.
Unlike typical static obstacles, an open door represents a non-rigid shape variation of a familiar car object, which can easily be overlooked or misinterpreted by models that lack fine-grained semantic understanding of the scene.
The ego vehicle controlled by $\mathcal{M}_{base}$ fails to recognize the shape variation.
It comes to a halt and subsequently becomes immobilized, unable to plan a feasible path around the obstruction.
LVDrive accurately perceives the altered vehicle shape and correctly interprets it as a navigational hazard, successfully planning a safe trajectory that bypasses the car with the open door and continues forward without interruption.
\textbf{This comparison highlights LVDrive's ability to capture fine-grained semantic details and reason about their implications for safe motion planning.}

\begin{figure*}[ht]
    \centering
    \includegraphics[width=0.95\textwidth, trim=0 30 0 0]{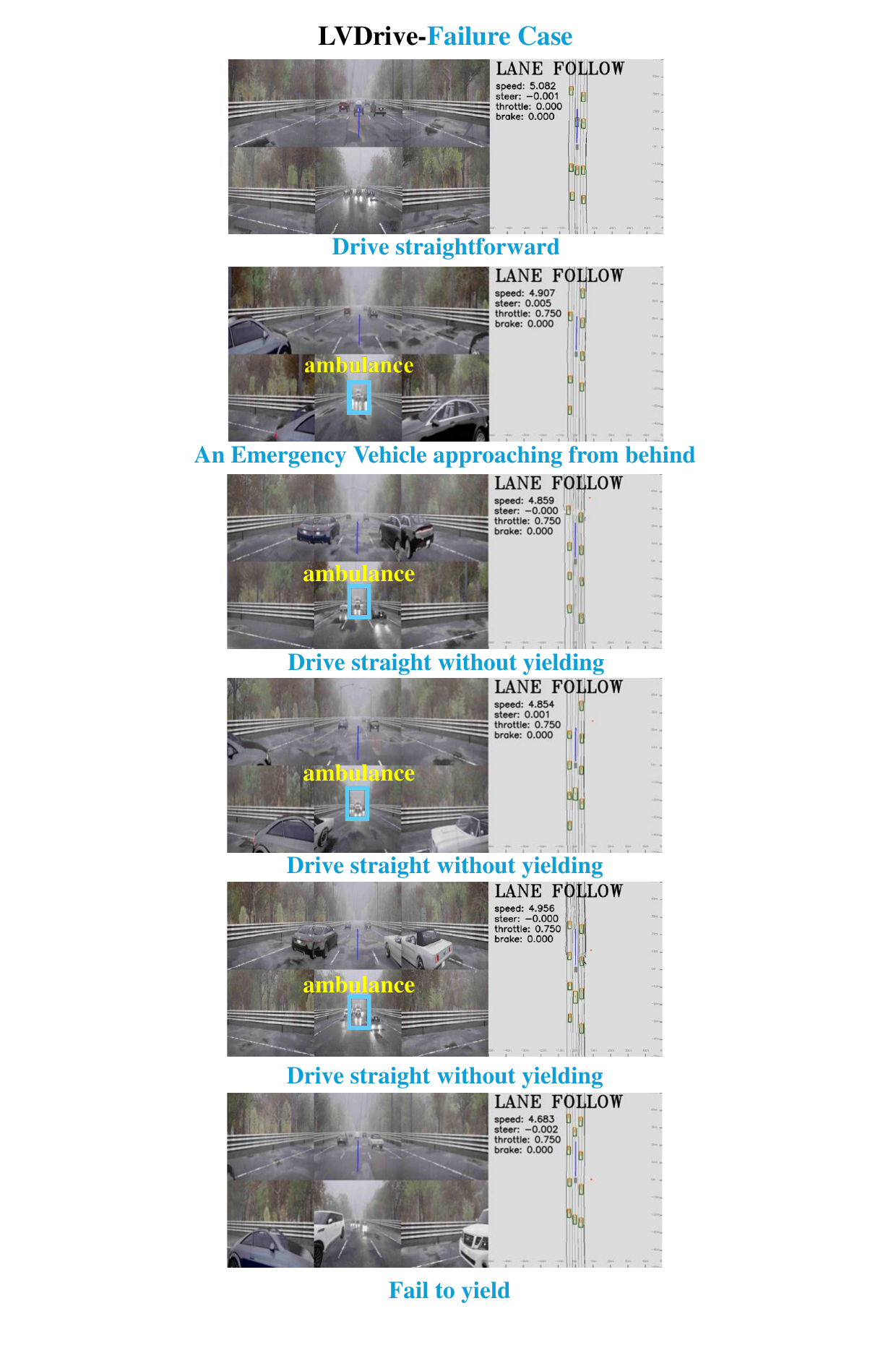}
    \caption{
    \textbf{Failure case of our LVDrive in a Give Way scenario.}
    The ego vehicle is required to yield to the emergency vehicle that approaches from behind.
    LVDrive maintains the straight route and fails to yield to the ambulance.}
    \label{fig:vis_a5}
\end{figure*}

\paragraph{Failure case on Give Way.}
One failure case in \textit{Give Way} scenario is shown in Figure~\ref{fig:vis_a5}.
The ego vehicle is required to yield to an emergency vehicle that approaches from behind.
However, LVDrive maintains its straight route and fails to yield to the ambulance.
We attribute this failure to the incorporation of the front-view future scene prediction task, which inherently leads the model to prioritize front-view scene representations.
As a result, LVDrive overlooks the rear interaction and fails to correctly identify the approaching vehicle as an emergency ambulance, leading to an unsuccessful Give Way task.


\clearpage



\end{document}